\pdfoutput=1

\documentclass[11pt]{article}

\usepackage[final]{acl}

\usepackage{times}
\usepackage{latexsym}

\usepackage[T1]{fontenc}

\usepackage[utf8]{inputenc}

\usepackage{microtype}
\usepackage{inconsolata}
\usepackage{graphicx}
\usepackage{amsmath}
\usepackage{listings}

\usepackage{booktabs}
\usepackage{tabularx}
\usepackage{arabtex}

\newcommand{\modelname}{SparQLe}

\title{\modelname: Speech Queries to Text Translation Through LLMs}

\author{Amirbek Djanibekov, Hanan Aldarmaki \\
  Mohamed bin Zayed University of Artificial Intelligence \\
Abu Dhabi, UAE \\
 \texttt{\fontsize{10}{12}\selectfont \{amirbek.djanibekov;hanan.aldarmaki\}@mbzuai.ac.ae}}

\begin{document}
\maketitle
\begin{abstract}

With the growing influence of Large Language Models (LLMs), there is increasing interest in integrating speech representations with them to enable more seamless multi-modal processing and speech understanding. This study introduces a novel approach that combines self-supervised speech representations with instruction-tuned LLMs for speech-to-text translation. The proposed approach leverages a modality adapter to align extracted speech features with instruction-tuned LLMs using English speech data. Our experiments demonstrate that this method effectively preserves the semantic content of the input speech and serves as an effective bridge between self-supervised speech models and instruction-tuned LLMs, offering a promising approach for various speech understanding applications.
\end{abstract}

\section{Introduction}

Progress in speech processing has been accelerated by the introduction of self-supervised learning (SSL) methods that utilize large amounts of unlabeled speech data, which established new benchmarks in the field~\cite{xu2021self,hsu2021hubert,zeghidour2021soundstream}. 
Continuous representations and/or discrete  units derived from self-supervised models have been used to extract relevant latent features from speech data and improve performance in downstream tasks, including speech recognition~\cite{baevski2020wav2vec}, speech synthesis~\cite{renfastspeech,wang2023neural}, speech translation~\cite{inaguma2020espnet} and general speech understanding~\cite{wang2020transformer}.
Progress in text processing has also been accelerated by the emergence of pre-trained Large Language Models (LLMs), which enabled new applications such as few-shot/zero-shot language processing~\cite{radford2019language, brown2020language, touvron2023llama, qwen} and multi-modal processing~\cite{tsimpoukelli2021multimodal, radford2021learning}. Recent efforts in speech understanding explored the possibility of incorporating speech representations directly into LLMs~\cite{zhang2023speechgpt, wang2023slm, das2024speechverse, fang2024llama}. 
Multi-modal speech-language models signify a shift in both speech and natural language processing. By incorporating speech data, LLMs can enhance their contextual grasp, providing a deeper and more thorough representation of spoken language. In addition, the existing multilingual functionalities of LLMs can be leveraged to enhance speech processing applications, such as speech translation, without additional dedicated training.    

In this work, we describe an efficient method to query instruction-tuned LLMs using speech input. The model aligns speech features extracted through self-supervised learning (SSL) with LLMs using only a modality adapter trained with English data and a small portion of translated text. We demonstrate the generalization of translation performance across both seen and unseen target languages. We call our approach \modelname~\footnote{\textbf{Sp}eech \textbf{R}outing to \textbf{Q}uery Large \textbf{L}anguage models.}. \modelname~is inspired by Querying Transformer modules used in vision language models to bootstrap vision-language representations from frozen image encoders~\cite{li2023blip}. We demonstrate through speech translation that \modelname~enables the integration of existing pre-trained speech encoders and LLMs without the need for updating the parameters of either speech encoder or LLM. 
In contrast to previously explored speech-LLM integration approaches, our method is the first to utilize frozen SSL speech representations, without relying on large pre-trained ASR models like Whisper~\cite{radford2023robust}. We experimentally demonstrate the effectiveness of this relatively simple approach and release both the pre-trained and fine-tuned models~\footnote{\url{https://github.com/djanibekov/rebooting-llm}}.

\begin{table*}[http]
    \centering
    \resizebox{2\columnwidth}{!}{
    \begin{tabular}{lcccc}
    \toprule
         Method & Speech Encoder (Param.) & Language Model (Param.) & Adapter (Param.) & Tasks \\
        \midrule
       ~\cite{chen2024salm} & NeMo (0.6B - 1.1B) & MegatronLLM (40B-1T) & LoRA (14M-94M) / Conformer (115M) & multitask\\
       ~\cite{wang2023blsp} & Whisper (74M-1.5B) & LLama (6.7B-65.2B) & Conv.Layers (4M) & alignment \\
       ~\cite{wang2023slm} & USM (2B) & mT0-MT XXL (13B) & Adapter (156M) & multitask \\
       ~\cite{wang2023speech} & CTC encoder (220M) & T5 XXL (11B) - RAG & Speech2Text, Speech2Entity Retriever & multitask\\
       ~\cite{yu2024connecting} & Whisper (1.5B) & VicunaLLM (13B) & FC\footnote{Forward Connection} (24M) / MHSA\footnote{Multi-Head Self Attention} (133M) / Seg-Q-Former (24M) & ASR \\
       ~\cite{tang2024salmonn} & Whisper (1.5B) + BEATS (90M) & VicunaLLM (13B) & LoRA + Seg-Q-Former (33M) & ASR/multitask \\
       ~\cite{das2024speechverse}   & WavLM (316.62M) & Flan-T5-XL (2.85B) & CNN + LoRA(14M-94M) & multitask \\
       ~\cite{Qwen2-Audio}          &  Whisperlarge (1.5B) & Qwen (7B) & --- & multitask \\
        \midrule
        \modelname~& HuBERT (316M) &  LLama3 (8B) & Q-Former (187M) & AST/multitask \\
        
        \bottomrule
    \end{tabular}
    }
    \caption{Comparison of related works and proposed model. LoRA's rank in~\cite{chen2024salm} is assumed to be 8. For other rank values, multiply number of parameters by 2 for 16 and 4 for 32 ranks, respectively.}
    \label{tab:related_works_comparison}
\end{table*}
\section{Related Works}

The availability  of instruction-tuned LLMs~\cite{touvron2023llama, llama3modelcard, jiang2023mistral} opened a new research direction for speech processing by connecting speech directly to these multi-task models. \citet{chen2024salm} proposed multitask speech-language modeling with unified LLM framework that shows in-context learning ability. \citet{yu2024connecting} utilized three approaches for adapting speech to text modality: Fully Connected Linear Layers following ~\cite{houlsby2019parameter} adapter method, multi-head cross attention mechanism described in~\cite{vaswani2017attention}, and query transformer~\cite{li2023blip}. For processing speech input, they utilized two models: Whisper Large-v2~\cite{radford2023robust} and BEATS~\cite{pmlr-v202-chen23ag}. The SpeechVerse~\cite{das2024speechverse} framework used WavLM-based~\cite{chen2022wavlm} speech encoder interfaced with a Flan-T5-XL~\cite{JMLR:v25:23-0870} language model.
In a another study, \citet{ma2024embarrassinglysimpleapproachllm} demonstrated the sufficiency of a single linear layer for speech-LLM integration in ASR, albeit with limited exploration beyond this task.
Speech as language modeling was also studied in SpeechGPT \citep{zhang2023speechgpt} which integrates both speech and text modalities. The model incorporates a speech tokenizer that converts raw audio waveforms into discrete speech tokens, enabling efficient processing within the transformer architecture. Through multi-task fine-tuning on downstream tasks such as ASR, translation, and generation, the model demonstrates remarkable versatility. Qwen2-Audio~\cite{Qwen2-Audio}, designed as a general-purpose audio understanding model, exhibiting broad applicability across various audio-related tasks. The model employs self-supervised learning techniques, such as masked audio modeling and contrastive learning, to capture rich audio representations.

Table \ref{tab:related_works_comparison} summarizes the features of most relevant related works. 
We outline that, depending on the rank of the LoRA~\cite{hu2021lora} adapter, the final number of trainable parameters can increase. LoRA rank is the number of linearly independent rows or columns in a parameter (weight) matrix; a lower rank means approximating a large weight matrix with fewer parameters to simplify and speed up training. The number of additional parameters can be roughly estimated as the initial hidden dimension multiplied by the rank and then multiplied by two to account for all added parameters. In Table~\ref{tab:related_works_comparison}, we outline the range of the possible numbers of added parameters. 
Note that our proposed model, \modelname, is the only one that relies exclusively on SSL features (i.e. HuBERT) as input, and a simple adapter between the frozen speech encoder and LLM; previous approaches relied on complex encoders that have already been aligned with text through supervised training or adapt selected LLM with LoRA adapter. 

\section{Model}

\modelname~is a parameter efficient model designed to extract information from speech representation and route them to query pre-trained open-sourced LLMs, without modifications to the underlying speech encoder or LLM. Motivated by the success of multi-modal representations in vision language modeling~\cite{li2023blip}, we propose the adoption of speech representations to LLMs for generative tasks, specifically Automatic Speech Translation~(AST). We pre-trained our model using English data first and fine-tuned with mix of English and French. 

\begin{figure}[h]
\centering
    \includegraphics[width=0.6\columnwidth]{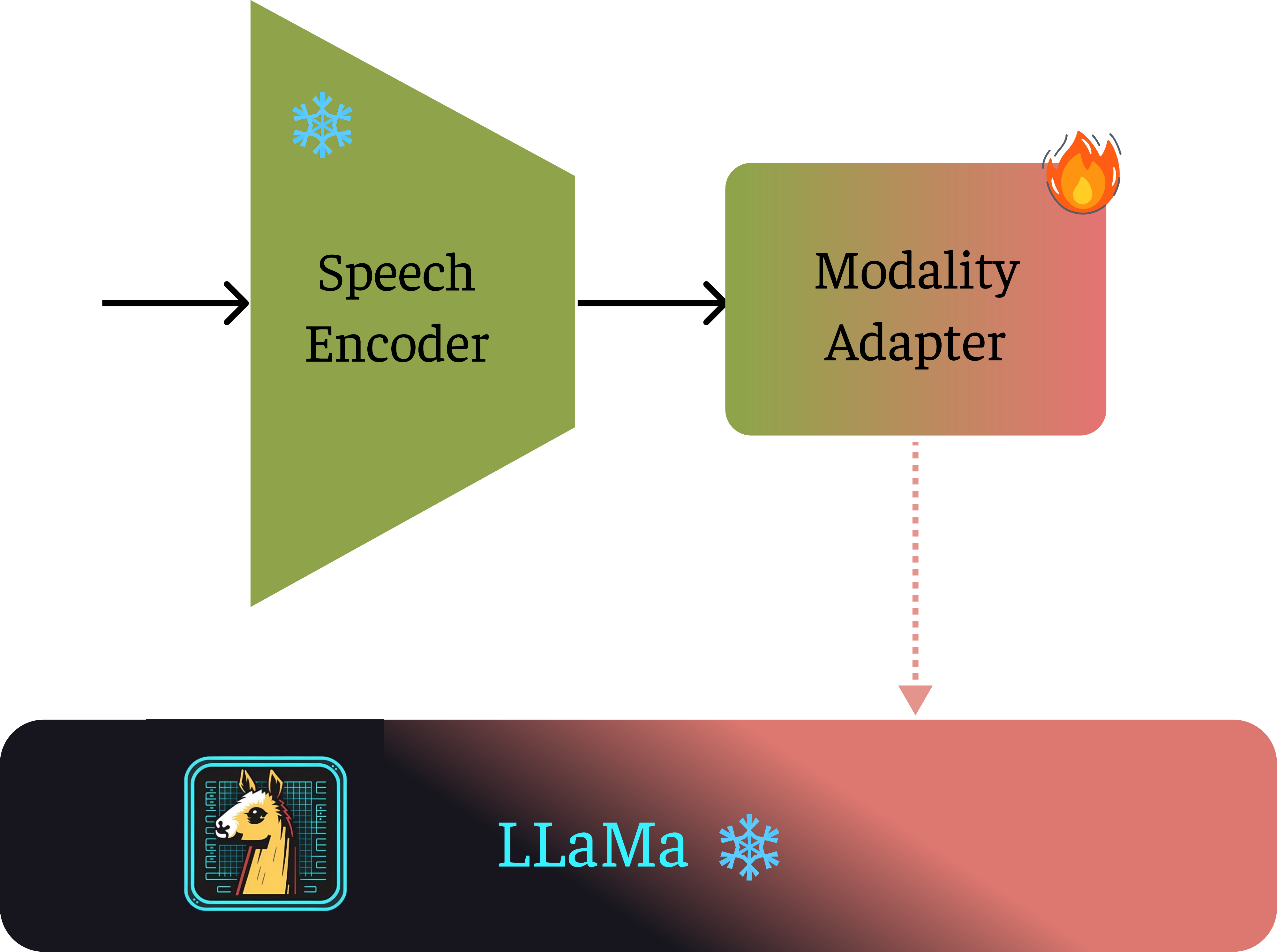}
    \caption{High-level overview of the \modelname~model}
    \label{fig:intermediate_a}
\end{figure}

We used HuBERT~\cite{hsu2021hubert} as the speech encoder. The output from its final hidden layer is fed into the query adapter. A query adapter incorporates query tokens, which are special tokens (placeholders) added to the input of a speech-language model. They do not correspond to specific speech regions, but are meant to extract information from the whole speech sequence in a flexible way. The final features from the query tokens are passed to a large language model to generate natural language responses. Figure~\ref{fig:intermediate_a} shows the overall structure of the system, which consists of three main parts: a pre-trained speech encoder, a bridging mechanism (\modelname) and a text generator. In our experiments, we employed LLama3~\cite{llama3modelcard} as the text language model.

\subsection{Pre-Training}
\begin{figure}
    \centering
    \includegraphics[width=\columnwidth]{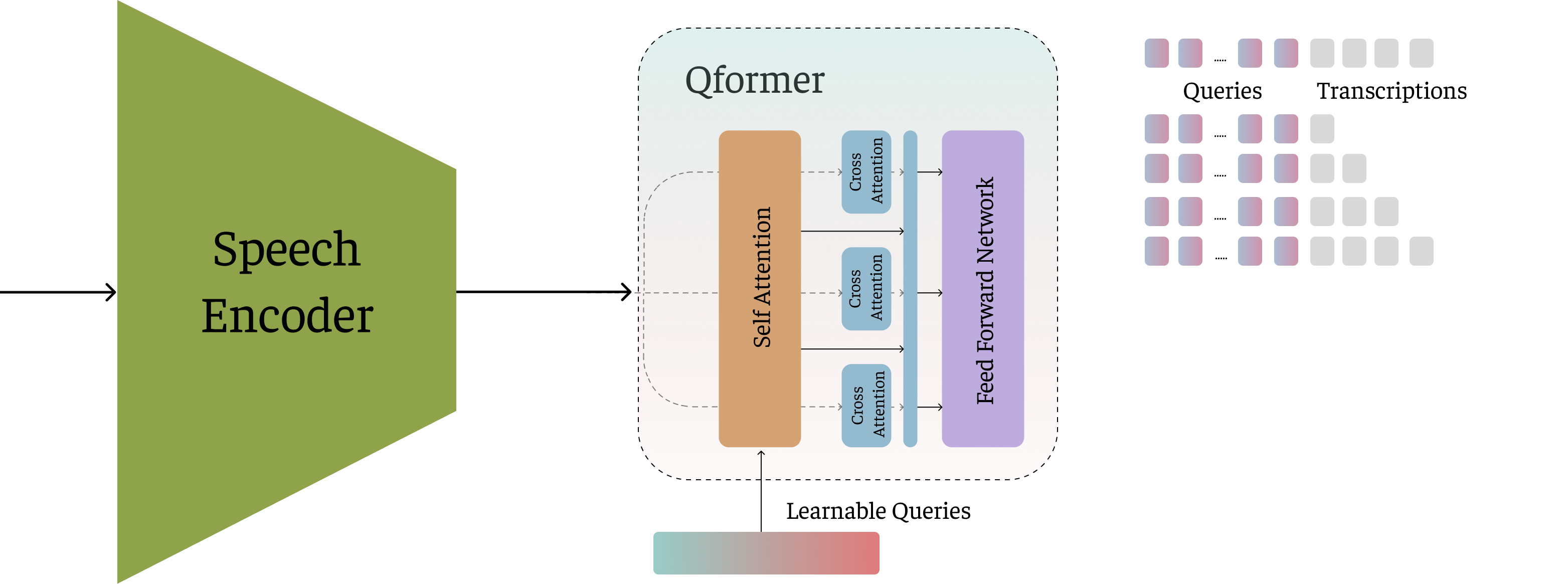}
    \caption{Modality adapter with auto-regressive supervised fine-tuning phase. The modality adapter is the Q-Former, which we discuss in the paper.}
    \label{fig:intermediate_b}
\end{figure}

This stage is akin to ASR training, where we utilize transcribed speech for supervised training. However, we do not introduce additional parameters and instead use the same modality adapter as an auto-regressive language model: each output vector from the Q-Former~\cite{li2023blip} is successively fed into a modality adapter to predict the next token. We only update the parameters of the adapter, and keep the underlying speech encoder frozen. 
The Q-Former is a vanilla transformer model but with learnable query tokens. These tokens are randomly initialized and designed to be learned during training to capture query information that is relevant to the task.
The process is depicted in Figure \ref{fig:intermediate_b}. 
In addition to the text generation task, we use various modality alignment objectives to account for speech in the input and aligned text-like features in the output, similar to image-text alignment done in \citet{li2023blip}: 
\textbf{Speech-text contrastive learning} 
aligns speech and text representation such that mutual information is maximized. This is achieved through contrasting speech-text cosine similarity of positive against negative pairs. 
\textbf{Speech-text matching} 
loss aligns representations of speech and text via a binary classification task. 
\textbf{Speech text generation}
loss trains the model to produce text based on the given audio. 

\subsection{Fine-tuning}
After pre-training, we fine-tune the adapter on downstream tasks using an instruction-tuned LLM, specifically LLama3. We utilize the extracted query tokens as the input to the LLM and update the adapter parameter using cross entropy loss derived from the LLM's objective. 
For instruction tuning, a frozen Large Language Model was fed with a randomly selected pool of prompts, which were designed to define the translation task. Subsequently, the instruction-tuned model was employed in a chat-based format to collect predictions.

\section{Experiments}

\begin{figure*}
    \centering
    \includegraphics[width=2\columnwidth]{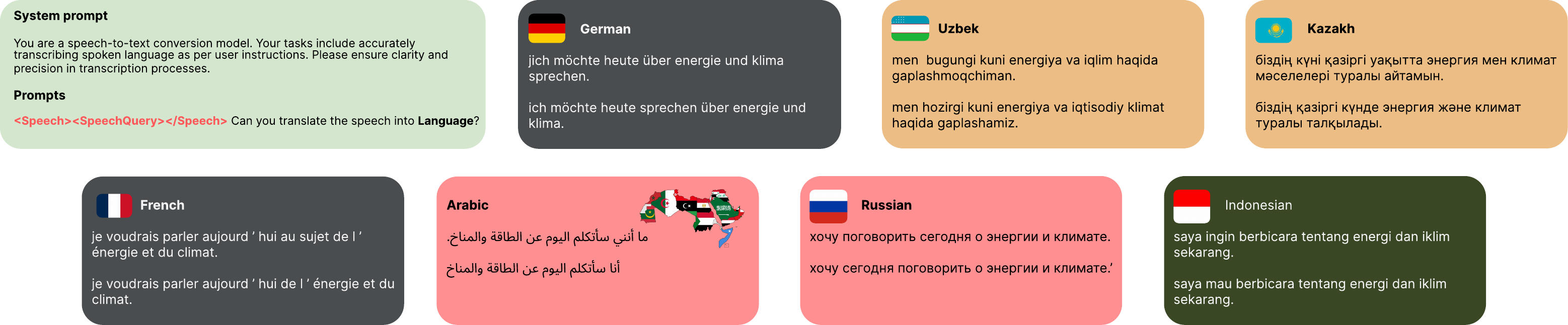}
    \caption{Sample of zero-shot instruction generation across multiple languages. To evaluate zero-shot capability, we simply changed the output language specified in the prompt. The produced text is lowercased and punctuation-free, following the text processing guidelines described in Section~\ref{sec:dataset_normalization}.}
    \label{fig:zeroshot_translation_examples}
\end{figure*}

\subsection{Datasets}
\label{sec:dataset_normalization}
To train and evaluate the model on Automatic Speech Translation (AST) task, we used the MuST-C~\cite{di-gangi-etal-2019-must} and LibriSpeech~\cite{panayotov2015librispeech} datasets. Specifically, we selected the French and German languages from MuST-C for AST evaluation.
We normalized the text across datasets by converting all letters to lowercase and eliminating punctuation marks. The MuST-C dataset includes action descriptions within the audio samples, such as "\texttt{<|speech|> (applause) <|speech|>}", which signify auditory sequences where spoken content is interspersed with audience applause. We opted to remove these actions from the translation text.

\subsection{Pre-Training}
\subsubsection{Experimental settings}
For feed-forward networks and self-attention of the modality adapter, we employ a 12-layer transformer-based Q-Former that is, by design, a UniLM~\cite{dong2019unified}; cross-attention is initiated randomly. 
For pre-training experiment, we trained the model using Adam optimizer, coupled with a cosine annealing learning rate scheduler during the pre-training. The learning rate was initiated at $1 \times 10^{-4}$ and gradually reduced to $1 \times 10^{-5}$, incorporating a warm-up phase at $1 \times 10^{-6}$. This means that learning rate started with warmup value and gradually reached from $10^{-6}$ to $10^{-4}$. The maximum length for speech samples was capped at 480K frames, which is equivalent to 30s of audio. We used 100 learnable query tokens in Q-Former.

Our experiments were conducted using open-source library for language-vision intelligence, LAVIS\footnote{\url{https://github.com/salesforce/LAVIS}}. 
The training processes were executed on one RTX4090 GPU with 24G memory being used over a period of two-three weeks with batch size equal to 8 due to memory constraints.

\subsection{Fine-Tuning}
\subsubsection{Experimental Settings}

We used 960 hours of audio from the LibriSpeech dataset, along with an additional 457 $\times$ 2 hours of audio samples from MuST-C that included both translation and transcription tasks: ~70\% of the speech samples for fine-tuning were used for recognition, while the remaining ~30\% involve English-to-French translation. We deliberately restricted our training data to one language in order to demonstrate the capacity of the model to generalize to other languages\footnote{Instruction tuning sometimes results in overfitting to the training instructions, as observed in \citet{tang2024salmonn}.}. 

\subsection{Prompts}
\label{sec:prompts}
We derived instruction prompts from SALMONN's \cite{tang2024salmonn} work. 
As demonstrated, each prompt includes a placeholder for speech \texttt{\textit{<Speech><SpeechQuery></Speech>}}, into which we insert query-extracted embeddings as inputs to the LLM. Please note that the query embeddings are placed inside the placeholder denoted by \texttt{\textit{<SpeechQuery>}}. Here is the prompts that we used for training:

\begin{itemize}
\setlength{\parskip}{0pt}
    \item  \texttt{<Speech><SpeechQuery></Speech>} Can you translate the speech into "Language"?
    \item \texttt{<Speech><SpeechQuery></Speech>} Please translate the speech you heard into "Language".
    \item \texttt{<Speech><SpeechQuery></Speech>} Listen to the speech and translate it into "Language".
    \item \texttt{<Speech><SpeechQuery></Speech>} Give me the Language translation of this "Language".
\end{itemize}
\texttt{"Language"} can be any language a user wants to add to instruction.

\subsubsection{Results \& Analysis}
We benchmarked translation against the IWSLT challenge baselines for speech-to-text translation using BERTScore~\cite{bert-score} as reported in~\cite{anastasopoulos2022findings}. The results for English-German translation are zero-shot since the model is only fine-tuned with English-French speech translation data. 
Evaluating LLM answers for speech translation is a challenging task, primarily due to the presence of chat-specific artifacts in the output, such as prompt repetition, follow-up comments, and connecting phrases (e.g., \texttt{"here is the transcribed text:"}). To address this issue, we implemented a post-hoc approach in which we endeavored to eliminate instances of prompt recurrence~(chat artifacts) in the final text.
\begin{table}[t]
\centering
    \resizebox{\columnwidth}{!}{
    \begin{tabular}{lccc}
    \toprule
    & \multicolumn{1}{c}{MuST-C\_{En-Fr}} & \multicolumn{1}{c}{MuST-C\_{En-De}} \\
    & BERTScore $\uparrow$ & BERTScore $\uparrow$ \\ 
    \midrule
    STRONGBASELINE & 81.75\% & 77.44\% \\
    WEAKBASELINE & 77.28\% & 74.86\%  \\
    \midrule
    \modelname~  &  \textbf{85.56\%} &  \textbf{83.26\%}  \\
    \bottomrule
    \end{tabular}
    }
    \caption{Comparison of \modelname~against strong and weak baselines from  the IWSLT isometric speech  challenge~\cite{anastasopoulos2022findings}.}
    \label{tab:instruction_translation_transcription_resutls}
\end{table}
We consider two baseline systems from IWSLT2022 campaign~\cite{anastasopoulos2022findings}: 
\textbf{WEAKBASELINE} refers to a standard neural machine translation model trained under limited data conditions, without incorporating any isometric translation features.
\textbf{STRONGBASELINE} is trained using unconstrained data setting and incorporates output length control following the approach of \citet{lakew2021machine}. This method involves adding a length token at the beginning of the input, generating N-best candidate translations, and then re-ranking them based on a weighted combination of the model’s score and the length ratio.

The results in Table~\ref{tab:instruction_translation_transcription_resutls} highlight the generalization potential of the model in translation. Specifically, the BERTScore for the tst-COMMON split in the French language demonstrates that our system has surpassed both the WEAKBASELINE and STRONGBASELINE in terms of semantic similarity. 
Furthermore, evaluations on the tst-COMMON split for the German language show that the performance quality  extends to languages not included in the training set. This success can be attributed to the inherent translation performance of the underlying LLM, demonstrating the model’s adaptability to new instructions.\footnote{During the inference phase, we executed four different prompts which are listed in Section~\ref{sec:prompts}. We selected the prompt that yielded the best results on held-out set.}.

\section{Discussion}

We introduced a framework for efficient routing of SSL speech features to query LLMs, and demonstrated its effectiveness in speech translation tasks. The results indicate that the proposed model and training paradigm result in generalized performance and avoid instruction over-fitting; the model was able to adhere to instructions for translating speech into multiple target languages (see Figure~\ref{fig:zeroshot_translation_examples}). \modelname~demonstrates ability to translate speech input into diverse languages not encountered during our fine-tuning stage, such as German, Russian, Arabic, etc. Finally, with the appropriate prompts, the instruction-tuned model is capable of performing multiple tasks in a single run, (see Figure~\ref{fig:multitask_example} in Section~\ref{sec:multitask_example}). The performance in speech translation shows promising results, where the proposed approach outperformed both weak and strong baselines from \citet{anastasopoulos2022findings} in both French and German. 

\subsection{Multi task Discussion}
\label{sec:multitask_example}

\begin{figure}[htt]
    \centering
    \includegraphics[width=0.9\columnwidth]{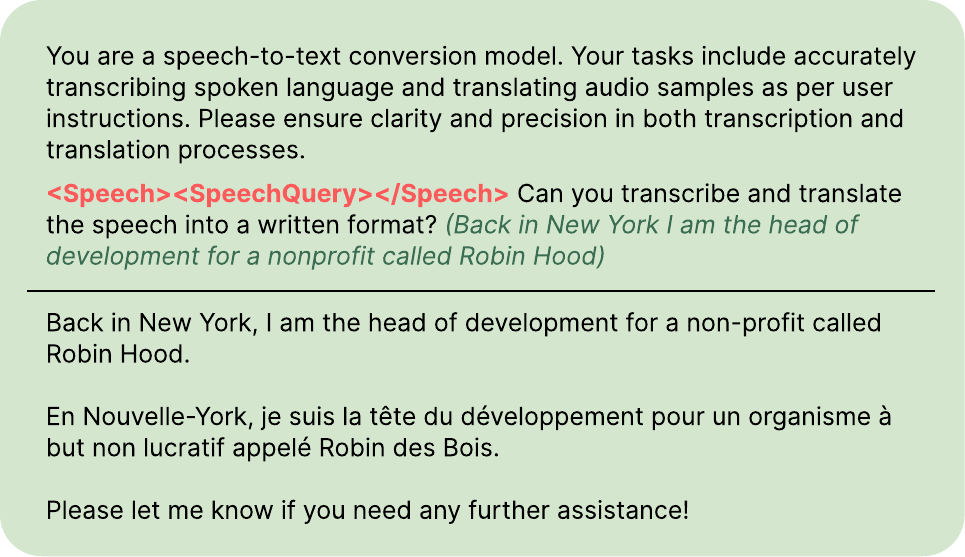}
    \caption{Example from the \modelname~for multi-tasking in one prompt.}
    \label{fig:multitask_example}
\end{figure}

As mentioned before with the appropriate prompts, the instruction-tuned model is capable of performing multiple tasks in a single run, See Figure~\ref{fig:multitask_example}. While we have not conducted an exhaustive analysis of this aspect in the current study, this example illustrates potential applications for efficiency and versatility in spoken language applications.

\section{Conclusion \& Future Work}
In this short paper, we demonstrate the performance of the proposed \modelname~model, an aligned speech-to-text model based on SSL features, for speech translation applications. What we have demonstrated in this study is only a subset of potential applications of this method. The \modelname~model can potentially handle both text and speech modalities, and can be applied for any speech-to-text applications. As demonstrated, our model outperforms existing speech translation baselines from IWSLT 2022 challenge, which demonstrates the potential of transferring the inherent capacities of LLMs into speech tasks using a parameter-efficient approach. Future work can explore the generalization of the model to other languages and speech understanding tasks and analyze the characteristics of the resulting queries. 
\clearpage
\section*{Limitations}
Our model was initially pre-trained to align specifically with English speech samples, disregarding other rich languages that present unique challenges. While we believe \modelname~has the potential to handle various tasks beyond its original training scope, we have not yet carried out a formal assessment to verify this capability. Although our model is adaptable to multiple LLMs, we onlye explored one model. Similarly, we did not explore other speech encoders apart from HuBERT.
For translation evaluation we used BERTScore, which measures semantic similarity for generation tasks, but all automatic translation metrics have limitations. For example, sentences \texttt{"never had any act seemed so impossible"} and \texttt{"always had any act seemed so impossible"} convey different information but are similar in words. BERTScore outputs that these two sentences have a high similarity score, which is, in fact, not true (99.7\% in F1 score).
We did not test our model on tasks other than translation and transcription. As a result, the model’s performance on other modalities or tasks, such as speech question answering, remains unverified.

\bibliography{custom}
\end{document}